\def\year{2020}\relax
\documentclass[letterpaper]{article} 
\usepackage{aaai20}  
\usepackage{times}  
\usepackage{helvet} 
\usepackage{courier}  
\usepackage[hyphens]{url}  
\usepackage{graphicx} 
\usepackage{graphicx}  
\usepackage{amsmath}
\usepackage{multirow}
\usepackage{threeparttable}
\usepackage{booktabs}

\title{Knowledge Graph Transfer Network for Few-Shot Recognition}
\author{
Riquan Chen,\textsuperscript{\rm 1}
Tianshui Chen,\textsuperscript{\rm 1,2}
Xiaolu Hui,\textsuperscript{\rm 1}
Hefeng Wu,\textsuperscript{\rm 1\,}{\thanks {Riquan Chen and Tianshui Chen contribute equally and share
first-authorship. Corresponding author is Hefeng Wu. This work is partly supported by National Natural Science Foundation of China (No. U1611461 and 61876045) and SenseTime Research Fund.}}
Guanbin Li,\textsuperscript{\rm 1}
Liang Lin\textsuperscript{\rm 1,2}
\\ 
\textsuperscript{\rm 1}Sun Yat-sen University,   
\textsuperscript{\rm 2}DarkMatter AI Research \\
sysucrq@gmail.com, tianshuichen@gmail.com,
huixlu@mail2.sys.edu.cn, wuhefeng@gmail.com \\
 liguanbin@mail.sysu.edu.cn, linlng@mail.sysu.edu.cn
}

\begin{document}

\maketitle

\begin{abstract}
Few-shot learning aims to learn novel categories from very few samples given some base categories with sufficient training samples. The main challenge of this task is the novel categories are prone to dominated by color, texture, shape of the object or background context (namely \textit{specificity}), which are distinct for the given few training samples but not common for the corresponding categories (see Figure \ref{fig:motivation}). Fortunately, we find that transferring information of the correlated based categories can help learn the novel concepts and thus avoid the novel concept being dominated by the \textit{specificity}. Besides, incorporating semantic correlations among different categories can effectively regularize this information transfer.  In this work, we represent the semantic correlations in the form of structured knowledge graph and integrate this graph into deep neural networks to promote few-shot learning by a novel Knowledge Graph Transfer Network (KGTN). Specifically, by initializing each node with the classifier weight of the corresponding category, a propagation mechanism is learned to adaptively propagate node message through the graph to explore node interaction and transfer classifier information of the base categories to those of the novel ones. Extensive experiments on the ImageNet dataset show significant performance improvement compared with current leading competitors. Furthermore, we construct an ImageNet-6K dataset that covers larger scale categories, i.e, 6,000 categories, and experiments on this dataset further demonstrate the effectiveness of our proposed model.
\end{abstract}

\section{Introduction}

\begin{figure}[!t]
   \centering
   \includegraphics[width=0.96\linewidth]{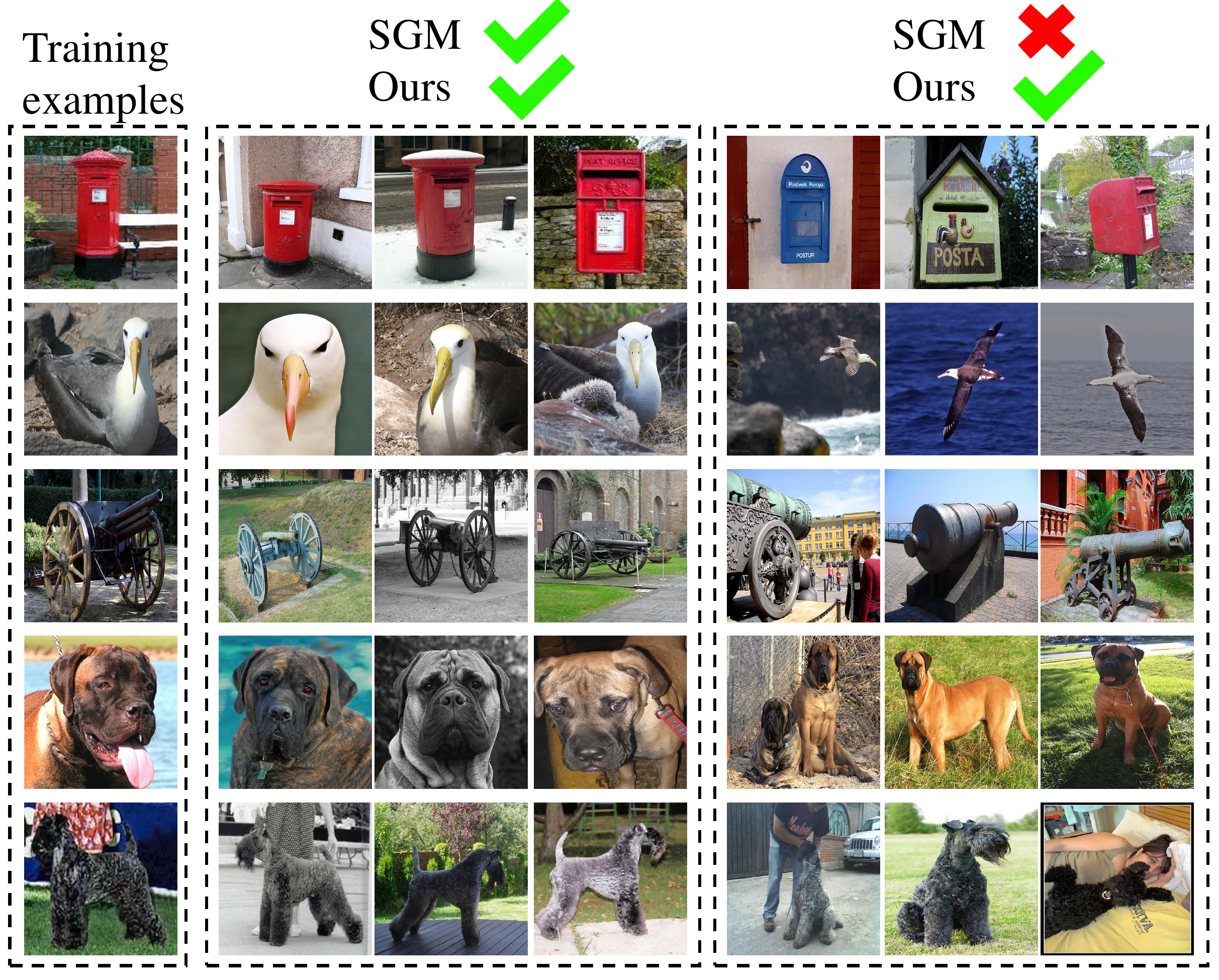}
   \caption{Visualization results of SGM and our proposed models. The first column shows the only training samples of different categories. The next three columns show the examples that are correctly classified by both two models. The last three columns show the samples that are misclassified by the SGM model but correctly classified by our model. It can be observed that the SGM model can well classify samples with high similarity to the training image but fails in the samples with a large difference in appearance. In contrast, our proposed model shows superior performance on more appearance patterns.}
    \label{fig:motivation}
\end{figure}

Recently, deep convolutional neural networks (ConvNet) \cite{krizhevsky2012imagenet,he2016deep} have obtained remarkable success in various visual recognition tasks. To fully train a deep ConvNet recognition system, it is requested that each category has thousands of training samples. If the system needs to recognize some novel categories, we need to collect large amounts of training samples for these categories to avoid being overfitting. In contrast, human can effortlessly learn to recognize novel categories with few samples by exploiting prior knowledge accumulated from daily life. Mimicking this ability to learn novel categories from very few samples (also known as few-shot learning) is a crucial yet practical task in the computer vision community.

Formally, we consider a general and practical scenario for few-shot learning, in which there exist a set of base categories with sufficient training samples and a set of novel categories with very limited samples (e.g., no more than ten samples in this work). Unlike previous works that merely focus on the novel categories, we aim to develop a recognition system that learns to recognize the novel categories based on few training samples and simultaneously maintain the performance on the base categories.

\begin{figure}[t]
   \centering
   \includegraphics[width=0.95\linewidth]{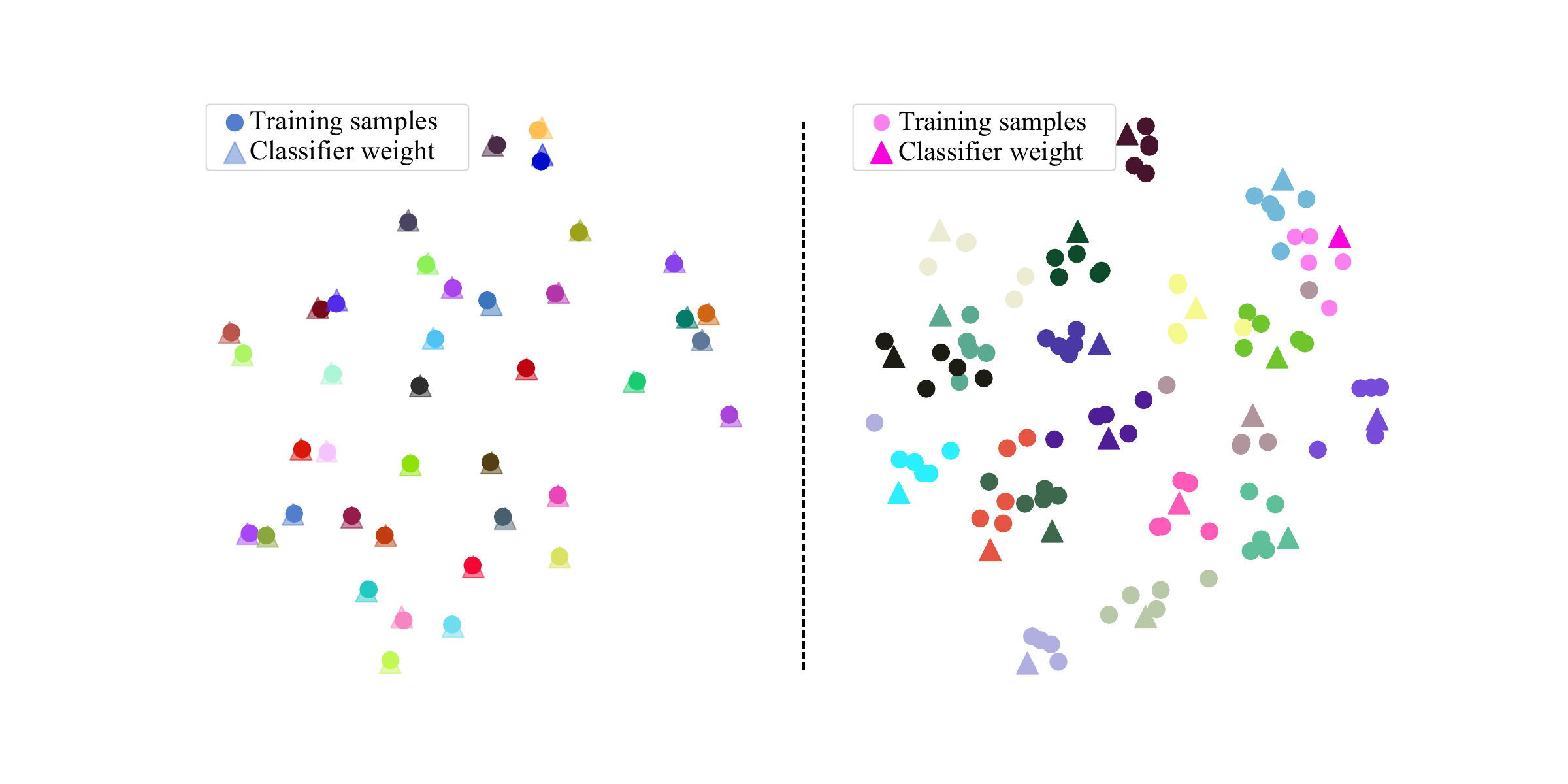}
   \caption{The t-SNE result of the normalized training samples and classifier weight. Scatter plot from the left to the right shows 1-shot and 5-shot setting, respectively.}
    \label{fig:motivation_classifier_weight}
\end{figure}

It is well-known that each image contains features that are particular and discriminant for this image but not common for its category (e.g., texture, color, shape of the object or background context). Here, we denote these features as \textit{specificity}. In few-shot learning scenarios, each novel category has very limited samples that can hardly describe its commonality, so its learned classifier is prone to dominated by the \textit{specificity} and thus may deviate severely from reality. Take the \textit{Kerry blue terrier} in the last row of Figure \ref{fig:motivation} for example, which is in the 1-shot setting on ImageNet-FS dataset, the state-of-the-art model SGM \cite{hariharan2017low} can well classify the instances with a specific side-standing pose but fail in instances with other poses. To delve deep into this phenomenon, we further visualize the extracted features of training samples and the corresponding classifier weights learned by SGM \cite{hariharan2017low} in Figure \ref{fig:motivation_classifier_weight}. As shown, the learned classifier is highly correlated with the extracted features. If the training samples of a category mainly contain \textit{specificity}, the learned classifier may squint towards describing this \textit{specificity} and inevitably miss the common feature for this category.

Fortunately, strong semantic correlations are witnessed among categories with similar visual concepts, which originates from human recognitive system. Transferring the information of the correlated base categories can provide additional information to guide learning the novel concept and thus avoid the novel concept being dominated by the \textit{specificity}. Moreover, these category correlations can effectively regularize this information transfer. In this work, we develop a Knowledge Graph Transfer Network (KGTN) that integrates category correlations into a deep neural network to guide exploiting information of base categories to help to learn the novel concept. To this end, we represent the semantic correlations with a knowledge graph, where each node refers to the classifier weight of a category and each edge represents the semantic correlation between the two corresponding categories. Then, we introduce a graph propagation mechanism to transfer node message through the graph. In this way, it allows each category to derive information of the correlated categories to better learn the classifier under the explicit guidance of category correlations. Notably, the graph contains both novel and base categories and the message propagation mechanism is shared across all node pairs in the graph, and thus such a mechanism can be trained using sufficient samples of the base categories and well generalizes to the novel categories.

The contributions are summarized into three folds: 1) We propose to integrate category correlations as prior knowledge to regularize and guide transferring information of classifier weights. 2) We introduce a graph update mechanism that propagates the node message through the graph to iteratively update the classifier weights. 3) We conduct experiments on the widely used ImageNet based few-shot dataset and demonstrate the superior performance of our proposed framework over existing state-of-the-art methods. To evaluate the performance on larger scale categories, we construct an ImageNet-6K dataset that covers 6,000 categories. Experiments conducted on this benchmark shows that our model still outperforms current leading methods.

\section{Related Works}

\noindent\textbf{Few-shot learning.}
Previous works \cite{finn2017model,li2017meta,vinyals2016matching,garcia2017few,snell2017prototypical} follow a simple $N$-way $K$-shot setting, where there are $N$ categories with $K$ training samples for each category, and $N$ is no more than 20. These works often adopt learning-to-learn paradigm that distills knowledge learned from training categories to help learn novel concepts. For example, \cite{vinyals2016matching,snell2017prototypical,sung2018learning,kim2019variational} learn an embedding and metric function from the base categories to well recognize samples in the novel categories.  Most of these works evaluate their algorithms on some small-scale datasets, e.g., miniImageNet with 64 base categories, 16 validation categories, 20 novel categories. Recently, some works \cite{hariharan2017low,chen2019image,hui2019graph} switch to a more general and practical setting, where the algorithms aim to recognize hundreds of novel concepts with very limited samples given a set of based categories with sufficient training samples. To address this few-shot learning scenario, \cite{hariharan2017low} learn a transformation function to hallucinate additional samples for novel categories. \cite{chen2019image} further consider to enlarge the size of dataset and learn a deformation network to deform images by fusing a pair of images. \cite{qiao2018few} explore the relation of training feature and classifier weight and adapt a neural network to obtain classifier weights directly from the training features. To evaluate these algorithms, researchers construct a larger-scale dataset that covers 1,000 categories \cite{hariharan2017low}. As this is a more practical scenario, we focus on this setting in our work.

\noindent\textbf{Knowledge Embedded Visual Reasoning.}
Recently, lots of works attempt to incorporate prior knowledge with deep representation learning for various visual reasoning tasks, ranging from visual recognition/detection \cite{wang2018zero,chen2018fine,lee2018multi,chen2019learning,chen2018knowledge,jiang2018hybrid} to visual relationship reasoning \cite{chen2019knowledge,wang2018deep} and navigation/planning \cite{WeiYang_ICLR2019,chen2018neural}. As a pioneering work, \cite{marino2017more} build a knowledge graph to correlate object categories and learn graph representation to enhance image representation learning. \cite{chen2019learning} incorporate a similar knowledge graph to guide feature decoupling and interaction to further facilitate multi-label image recognition.  \cite{wang2018zero} apply graph convolutional network \cite{kipf2016semi} to explore semantic interaction and direct map semantic vectors to classifier parameters. To further explore high-level tasks, \cite{chen2019knowledge} consider the correlations between specific object pairs and their corresponding relationships to regularize scene graph generation and thus alleviate the effect of the uneven distribution issue. \cite{chen2018neural} build And-Or Graphs \cite{zhu2007stochastic} to describe tasks, which helps regularize atomic action sequence generation to reduce the dependency on annotated samples. In few-shot learning scenario, the latest work \cite{li2019large} construct category hierarchy by semantic cluster and regularize prediction at each hierarchy via a hierarchical output net. Different from these works, we formulate classifier weight as a prototype representation of the corresponding category, and introduce a graph propagation mechanism to transfer prototype representation of base categories to guide learning novel concepts under the explicit guidance of prior semantic correlation.

\begin{figure*}[!t]
   \centering
   \includegraphics[width=0.9\linewidth]{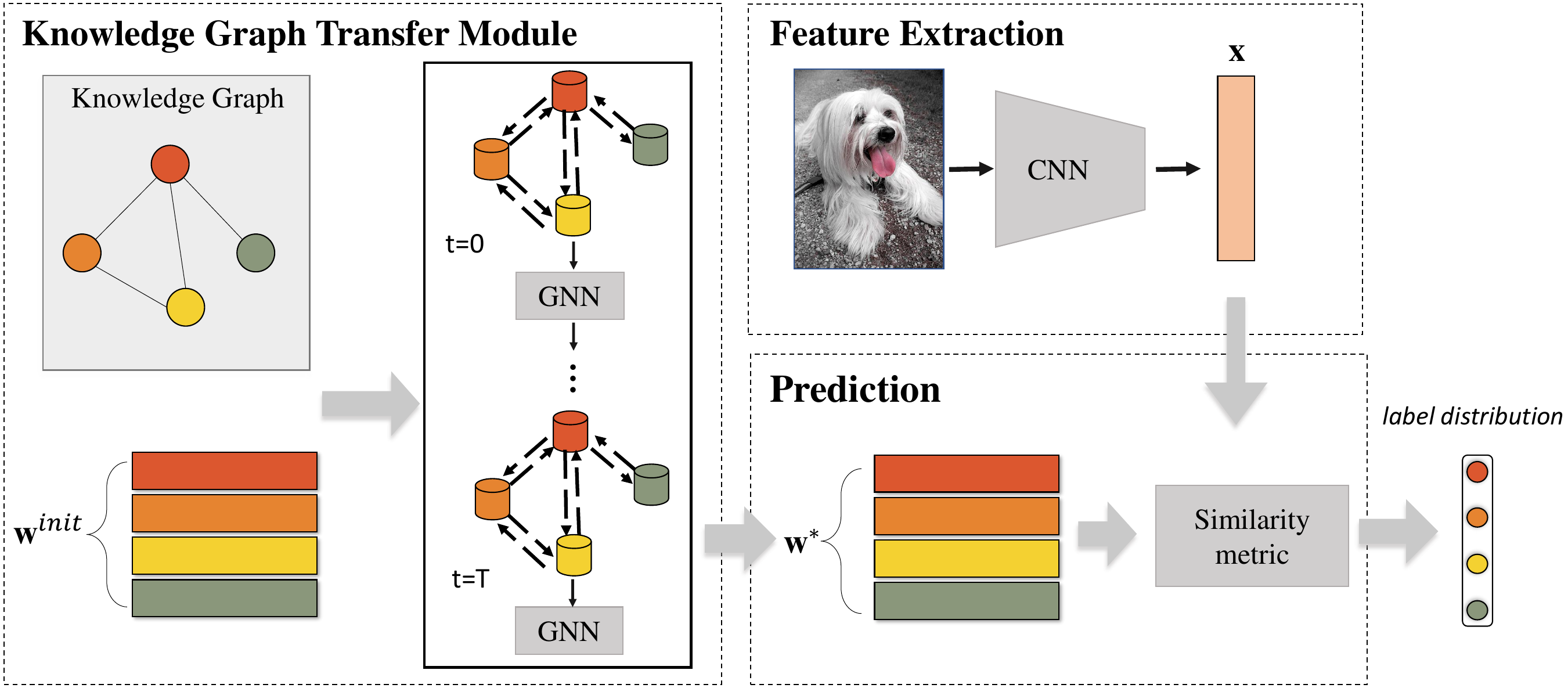} 
   \caption{Illustration of the proposed Knowledge Graph Transfer Network (KGTN) model. It incorporates prior knowledge of category correlations to help explore the interactions among classifier weights of all categories to better learn the classifier weights of the novel categories.}
   \label{fig:framework}
\end{figure*}

\section{Methodology}

We first revisit the few-shot problem formulation in the conventional deep ConvNet paradigm with some formalized interpretation on why such paradigm easily fails in the few-shot scenarios, and then present our model in detail.

\subsection{Problem Formulation Revisited}

A few-shot recognition system should be able to recognize novel categories with limited few samples.
Recent deep ConvNet models have achieved remarkable success in image recognition, but they are also notorious for requiring a large number of training samples for each category and are not fit for the few-shot scenarios.
In this work, we amend the conventional paradigm to adapt deep ConvNet models to such scenarios.

A typical deep ConvNet-based image recognition model consists of the feature extractor and classifiers, which are jointly learned for optimal performance. Given an image sample $x$, the recognition model predicts its label $\hat{y}$ as $\hat{y} = \mathop{\arg\max}_{k} p(y=k|x)$, and
\begin{equation}\label{eq:softmax}
p(y=k|x)=\frac{\exp(f_k(\mathbf{x}))}{\sum_{i=1}^K\exp(f_i(\mathbf{x}))}
\end{equation}
is the softmax function. Here $K$ is the number of categories, and $\mathbf{x}$ is the $d$-dimensional representation feature of $x$ output by the feature extractor $\phi(\cdot)$, i.e., $\mathbf{x}=\phi(x)$. The linear classifier $f_k(\mathbf{x})= \mathbf{w}_k^{\top}\mathbf{x}+b_k$ computes the confidence of the sample $x$ belonging to category $k$, and is implemented with a fully-connected layer. $\mathbf{w}_k$ denotes the classifier weight and $b_k$ is the bias term. It can be easily inferred that $\mathop{\arg\max}_{k} p(y=k|x)$ is equivalent to $\mathop{\arg\max}_{k} f_k(\mathbf{x})$.

We reformulate $f_k(\mathbf{x})$ as follows:
\begin{equation} \label{eq:dot_to_dis_prove}
   \begin{split}
    f_k(\mathbf{x}) & = \mathbf{w}_k^{\top}\mathbf{x}+b_k \\
		  &= - \frac{1}{2}\|\mathbf{w}_k - \mathbf{x}\|_2^2 + \frac{1}{2}\|\mathbf{w}_k\|_2^2 + \frac{1}{2}\|\mathbf{x}\|_2^2+ b_k
   \end{split}
\end{equation}
By introducing the constraints $b_k = 0 $ and $\|\mathbf{w}_i\|_2=\|\mathbf{w}_j\|_2$, $\forall i,j$, the classifier $f_k(\mathbf{x})$ can be viewed as a similarity measure between $\mathbf{x}$ and $\mathbf{w}_k$, and we have
\begin{equation}\label{eq:dot_to_dis_conclusion}
	\hat{y} = \mathop{\arg\max}_{k} f_k(\mathbf{x}) = \mathop{\arg\min}_{k} \|\mathbf{w}_k - \mathbf{x}\|_2^2
\end{equation}

Therefore, the weight $\mathbf{w}_k$ can be viewed as the prototype representation of category $k$, and the sample $x$ is predicted as category $k$ if its  feature $\mathbf{x}$ has the maximum similarity (or minimum distance) with prototype $\mathbf{w}_k$. In this perspective, we can implement $f_k(\mathbf{x})$ with different similarity metrics.

Relaxing the above constraints to general deep ConvNet models, the prototype representation perspective of classifier weight is reasonable to some extent. Thus, when these models are applied in the few-shot scenarios, the learned prototype will be guided to reflect the \textit{specificity} of the few training samples (as visualized in Figure \ref{fig:motivation_classifier_weight}) and cannot capture the commonality of the corresponding category.

To tackle this problem, we amend the conventional deep ConvNet paradigm by incorporating category correlations to transfer prototype presentations among similar categories and thus substantially enhance the prototypes of novel categories. Specifically, the category correlations are modeled as a knowledge graph and integrated into the deep ConvNet to build our Knowledge Graph Transfer Network model, which will be detailed below.

\subsection{Knowledge Graph Transfer Network}
The overall framework of our Knowledge Graph Transfer Network (KGTN) is illustrated in Figure \ref{fig:framework}, which consists of three modules: Feature Extraction, Knowledge Graph Transfer, and Prediction. The key design is the knowledge graph transfer module where we incorporate a graph neural network to explore the knowledge transfer of the prototypes (classifier weights) by the guidance of semantic correlations on top of the ConvNet.

\noindent\textbf{Knowledge Graph Transfer Module. } We model the classifier weight into a graph, in which nodes refer to the classifier weigh and edges represent their semantic correlations. Then, we incorporate a Gated Graph Neural Network (GGNN) to update and propagate the message between nodes.

Given a dateset that covers $K=K_{base}+K_{novel}$ categories ($K_{base}$ and $K_{novel}$ denote the number of base and novel categories), we use a graph $\mathcal{G}=\{\mathbf{V}, \mathbf{A}\}$ to encode the correlations among all categories, in which nodes refer to the categories and edges represent their correlations. Specifically, $\mathbf{V}$ is represented as $\{v_1, v_2, \dots, v_{K_{base}}, \dots, v_K\}$ where node $v_k$ corresponds to category $k$, and $\mathbf{A}$ is an adjacency matrix, in which $a_{ij}$ denotes the correlation weight between categories $i$ and $j$.

At each iteration $t$, node $v_k$ has a hidden state $\mathbf{h}_k^t$, and the hidden state $\mathbf{h}_k^0$ at iteration $t=0$ is set as the initial classifier weight $\mathbf{w}_{k}^{init}$, formulated as
\begin{equation}
\mathbf{h}_k^0=\mathbf{w}_{k}^{init},
   \label{eq:node-initialization}
\end{equation}
in which $\mathbf{w}_k^{init}$ is learnable parameters and randomly initialized before training. At each iteration $t$, each node $k$ aggregates message from its correlated node such that the parameter vectors of these nodes can help refine its parameter vector, formulated as
\begin{equation}
   \mathbf{a}_k^t=[\sum_{k'=1}^K{a_{kk'}\mathbf{h}_{k'}^{t-1}},\sum_{k'=1}^K{a_{k'k}\mathbf{h}_{k'}^{t-1}}].
   \label{eq:node-aggregation}
\end{equation}
In this way, a high correlation between nodes $k$ and $k'$ encourage message propagation from $k'$ to $k$, and it suppresses the propagation otherwise. Then, the framework takes this aggregated feature vector and the hidden state of the previous iteration as input to update the corresponding hidden state by a gating mechanism
\begin{equation}
   \begin{split}
    \mathbf{z}_k^t=&{}\sigma(\mathbf{W}^z{\mathbf{a}_k^t}+\mathbf{U}^z{\mathbf{h}_k^{t-1}}) \\
    \mathbf{r}_k^t=&{}\sigma(\mathbf{W}^r{\mathbf{a}_k^t}+\mathbf{U}^r{\mathbf{h}_k^{t-1}}) \\
    \widetilde{\mathbf{h}_k^t}=&{}\tanh\left(\mathbf{W}{\mathbf{a}_k^t}+\mathbf{U}({\mathbf{r}_k^t}\odot{\mathbf{h}_k^{t-1}})\right) \\
    \mathbf{h}_k^t=&{}(1-{\mathbf{z}_k^t}) \odot{\mathbf{h}_k^{t-1}}+{\mathbf{z}_k^t}\odot{\widetilde{\mathbf{h}_k^t}}
   \end{split}
   \label{eq:ggnn}
\end{equation}
where $\sigma(\cdot)$ and $\tanh(\cdot)$ are the logistic sigmoid and hyperbolic tangent functions, and $\odot$ is the element-wise multiplication operation. In this way, the model tends to adopt the more correlated message to update parameters of the current node. The propagation is repeated by $T$ iterations, and we can obtain the final hidden states, i.e., $\{\mathbf{h}_1^T, \mathbf{h}_2^T, \dots, \mathbf{h}_K^T\}$. Finally, we utilize a simple output network to predict the classifier weight
\begin{equation}
   \mathbf{w}_k^*=o(\mathbf{h}_k^T, \mathbf{h}_k^0).
   \label{eq:node-output}
\end{equation}

\noindent\textbf{Prediction with different similarity metrics. } As stated in the problem formulation, the classifier $f_k(\mathbf{x})$ can be implemented as similarity metric.  Here, we consider three similarity metrics for evaluation: \textit{inner product}, \textit{cosine similarity}, and \textit{Pearson correlation coefficient}.

\noindent\textbf{Inner product:}
\begin{equation}
   f_k(\mathbf{x})  = \mathbf{x} \cdot \mathbf{w}_k^* \\
   \label{eq:inner_product}
\end{equation}

\noindent\textbf{Cosine similarity:}
\begin{equation}\label{eq:Cosine_similarity}
   f_k(\mathbf{x})  =\frac{\mathbf{x} \cdot \mathbf{w}_k^*}{\|\mathbf{x}\|_2\cdot\|\mathbf{w}_k^*\|_2}
\end{equation}

\noindent\textbf{Pearson correlation coefficient:}
\begin{equation}
   f_k(\mathbf{x}) = \frac{(\mathbf{x}-\bar{\mathbf{x}}) \cdot (\mathbf{w}_k^* - \bar{\mathbf{w}}_k^*)}{\|\mathbf{x} - \bar{\mathbf{x}}\|_2\cdot\|\mathbf{w}_k^* - \bar{\mathbf{w}}_k^*\|_2} \\
   \label{eq:pearson_correlation}
\end{equation}
where $\bar{\mathbf{x}}$ is a $d$-dimensional vector with all elements being the same value computed by averaging all elements in $\mathbf{x}$, as is likewise for $\bar{\mathbf{w}}_k^*$.

The above-defined similarity metric will be put into the softmax function as denoted by Equation (\ref{eq:softmax}) to obtain the final prediction. For cosine similarity and Pearson correlation metrics, the values output by the softmax may be extremely small, since $\|f_k(\mathbf{x})\|_2 \le 1$ in such situations. So, similar to \cite{gidaris2018dynamic,qi2018low}, we multiply a learnable scale parameter $s$, i.e., putting $s\cdot f_k(\mathbf{x})$ into the softmax for these two metrics.

\subsection{Optimization}
Similar to \cite{hariharan2017low}, we adopt a two-stage training procedure to train the proposed model.

\noindent\textbf{Stage 1: } At the first stage, we train the feature extractor $\phi(\cdot)$ based on the base set $\mathcal{D}_{base}$. Given an image sample $x_i$ with label $y_i$, we first compute its probability distribution $\mathbf{p}_{i}=\{p_i^1, p_i^2, \dots, p_i^{k_{base}}\}$ with $p_i^k=p(y=k|x_i)$, and then define the cross-entropy loss as our objective function. To make the learned features easily generalize to the novel categories, we further introduce Squared Gradient Magnitude loss proposed in \cite{hariharan2017low} to regularize representation learning. Thus, the objective function at this stage can be defined as
\begin{equation}
   \mathcal{L}_1=\mathcal{L}_{c}+\lambda \mathcal{L}_{s}
   \label{eq:loss-1}
\end{equation}
where
\begin{equation}
   \begin{split}
    \mathcal{L}_{c}&=-\frac{1}{N_{base}}\sum_{i=1}^{N_{base}}\sum_{k=1}^{K_{base}}\textbf{1}(k=y_i)\log{p_i^k} \\
    \mathcal{L}_{s}&=\frac{1}{N_{base}}\sum_{i=1}^{N_{base}}\sum_{k=1}^{K_{base}}(p_{i}^{k}-\mathbf{1}(k=y_i))^2||\mathbf{x}_{i}||_2^2
    \end{split}
   \label{eq:loss-terms}
\end{equation}
where $N_{base}$ is the number of training samples in all base categories, $\textbf{1}(k=y_i)$ is an indicator that equals 1 when $k=y_i$ is true and 0 otherwise, $\lambda$ is a parameter to balance two loss terms and it is set as 0.005 by following \cite{hariharan2017low}. At this stage, the model is trained using the SGD algorithm with a batch size of 256, momentum of 0.9 and weight decay of 0.0005. The learning rate is initialized as 0.1 and is divided by 10 for every 30 epochs.

\noindent\textbf{Stage 2: } We fix the parameters of the feature extractor and use both the base and novel set to train the other components, including $\mathbf{w}^{init}$, the parameters of GGNN and scale factor $s$. Similarly, we obtain its probability vector $\mathbf{p}_{i}=\{p_i^1, p_i^2, \dots, p_i^{K}\}$ and use the cross-entropy loss as the objective function.
To handle class imbalance, we sample the novel and base samples by 1:1 ratio in each batch. Besides, for inner product, we introduce an additional regularization term on the classifier weighs, thus the overall objective function can be defined as

\begin{equation}
   \mathcal{L}_2=-\frac{1}{N}\sum_{i=1}^N\sum_{k=1}^{K}\textbf{1}(k=y_i)\log{p_i^k}+\eta\sum_{k=1}^K||\mathbf{w}_k^*||_2^2
   \label{eq:loss-2}
\end{equation}
where $N$ is the number of all training samples, $\eta$ is a parameter to balance two loss terms and it is set as 0.001 empirically. In this stage, we train the model using SGD algorithm with a batch size of 1,000, momentum of 0.9, weight decay of 0.0001, and learning rate of 0.01.

\renewcommand{\arraystretch}{1.2}
\begin{table*}[t]

  \centering
  \fontsize{8}{8}\selectfont
    \begin{tabular}{l|l|ccccccccc}
    \hline
    \multirow{2}{*}{Dataset}&
    \multirow{2}{*}{Method}&
    \multicolumn{4}{c}{Novel}&\multicolumn{1}{c}{ }&\multicolumn{4}{c}{All}\cr
    \cline{3-6} \cline{8-11}
    &&1&2&5&10& &1&2&5&10\cr
    \hline
    \multirow{10}{*}{\textbf{ImageNet-FS}}
    &MN\cite{vinyals2016matching}          &53.5      &63.5       &72.7       &77.4       & &64.9       &71.0       &77.0       &80.2\cr
    &PN\cite{snell2017prototypical}          &49.6      &64.0       &74.4       &78.1       & &61.4       &71.4       &78.0       &80.0\cr
    &SGM\cite{hariharan2017low}      &54.3      &67.0       &77.4       &81.9       & &60.7       &71.6       &80.2       &\bf83.6\cr
    &SGM w/ G\cite{hariharan2017low} &52.9      &64.9       &77.3       &82.0       & &63.9       &71.9       &80.2       &{\bf{83.6}}\cr
    &AWG\cite{gidaris2018dynamic}         &53.9      &65.5       &75.9       &80.3       & &65.1       &72.3       &79.1       &82.1\cr
    &PMN\cite{wang2018low}         &53.3      &65.2       &75.9       &80.1       & &64.8       &72.1       &78.8       &81.7\cr
    &PMN w/ G\cite{wang2018low}    &54.7      &66.8       &77.4       &81.4       & &65.7       &73.5       &80.2       &82.8\cr
    &LSD\cite{douze2018low}         &57.7      &66.9       &73.8       &77.6       & &--       &--       &--       &--\cr
    &KTCH\cite{li2019large}        &58.1      &67.3       &77.6       &81.8       & &--       &--       &--       &--\cr
    &IDeMe-Net\cite{chen2019image}   &60.1      &69.6       &77.4       &80.2       & &--       &--       &--       &--\cr
    \cline{2-11}
    &Ours(CosSim) &61.4 &70.4   &\textit{78.4} &\textit{82.2} & &\textit{67.7} &\textit{74.7} &\bf80.9 &\bf83.6 \cr
    &Ours(PearsonCorr) &\textit{61.5} &\textit{70.6}   &\bf78.5 &\bf82.3 & &67.5 &74.4 &80.7 &\textit{83.5} \cr
    &Ours(InnerProduct)    &{\bf{62.1}}&{\bf{70.9}}&{\textit{78.4}}&{\bf{82.3}}& &{\bf{68.3}}&{\bf{75.2}}&{\textit{80.8}}&\textit{83.5}\cr
     \hline
    \hline
    \multirow{4}{*}{\textbf{ImageNet-6K}}
 &SGM\cite{hariharan2017low}    &26.7&35.0&44.7&51.5& &37.7&44.8&53.1&58.5\cr
    &SGM w/ G\cite{hariharan2017low}&26.3&35.5&\bf{46.2}&\textit{52.0}& &39.4&46.4&\bf54.4&\bf{58.8}\cr
    &AWG\cite{gidaris2018dynamic}    &27.6&35.9&45.0&49.4& &39.3&45.2&52.1&55.4\cr
    \cline{2-11}
    &Ours(CosSim)    &\bf30.5&\bf37.5&46.0&51.7 & &\textit{41.0}&\textit{46.8}&\textit{53.8}&58.5\cr
    &Ours(PearsonCorr)   &\textit{30.4}&\textit{37.2}&45.9&51.8 & &\bf41.1&\textit{46.8}&53.7&58.5\cr
    &Ours(InnerProduct)    &29.5 &36.8&\textit{46.1}&\bf{52.1}& &40.8&\bf{46.9}&\bf{54.4}&\textit{58.7}\cr
    \hline

    \end{tabular}
    \vspace{2pt}
    \caption{Top5-accuracy in ``novel'' and ``all'' metrics of our model and current state-of-the-art methods on the ImageNet-FS and ImageNet-6K datasets. For fair comparison, all the methods use ResNet-50 for feature extraction. Some methods train an additional generator to hallucinate extra training samples for the novel categories (w/ G). The best and second best results are highlighted in {\bf{bold}} and {\textit{italic}}, respectively. ``-'' denotes the corresponding result is not provided.}
  \label{table:imagenet-sota}
\end{table*}

\section{Experiment}

\subsection{Graph Construction}
\label{sec:Graph Construction}
The knowledge graph encodes the correlations among different categories. It can be constructed according to different prior knowledge. Here, we introduce two kinds of knowledge, i.e., semantic similarity and category hierarchy.

\noindent\textbf{Semantic similarity. }Semantic word of a specific category well carries its semantic information, and the semantic distance of two categories encodes their correlations. In other words, two categories are of high correlation if their semantic distance is small, and are of low correlation otherwise. Thus, we first exploit this property to construct the graph. Specifically, given two categories $i$ and $j$ with semantic words $w_i$ and $w_j$, we first extract their semantic feature vector $\mathbf{f}_{i}^w$ and $\mathbf{f}_{j}^w$ using the GloVe model \cite{pennington2014glove} and compute their Euclidean distance $d_{ij}$. Then, we apply a monotone decreasing function $a_{ij}=\lambda^{d_{ij} - \mathop{\min}d_{ik},\forall k\ne i}$ (we set $\lambda=$0.4) to map the distance to the correlation coefficient $a_{ij}$. Besides, we also consider the self-message of each node so that we set $a_{ii}=1$.

\noindent\textbf{Category hierarchy. }Category hierarchy encodes category correlations via different levels of concept abstraction. Generally, the distance from one category to another indicates their correlation, where a small distance indicates a high correlation while a large distance indicates a low correlation. In this work, we also exploit this knowledge based on the WordNet \cite{miller1995wordnet} to construct the graph. Concretely, given two categories $i$ and $j$, we compute the shortest path from node $i$ to $j$ as the distance $d_{ij}$ and we also apply a similar monotone decreasing function to map the distance to the correlation coefficient $a_{ij}$.

\subsection{Datasets}

Unlike previous few-shot benchmark with low-resolution images and few novel categories, we consider the more realistic dataset with large scale base and novel categories: \emph{ImageNet Few-Shot (ImageNet-FS)} dataset. To further verify our approach, we construct a more challenging dataset \emph{ImageNet-6K}, which covers 5,000 novel categories.

\noindent\textbf{ImageNet-FS. }In this work, we first evaluate our proposed model on the widely used ImageNet-FS dataset. The dataset covers 1,000 categories from ILSVRC2012 and is divided into 389 base categories and 611 novel categories where 193 base categories and 300 novel categories are used for cross-validation and the remaining 196 categories and 311 novel categories are used for testing. Each base category has about 1,280 training images and 50 test images.

\noindent\textbf{ImageNet-6K. }To evaluate our proposed method on larger scale categories, we further construct a new benchmark ImageNet-6K that covers 6,000 categories. It contains 1,000 categories from ImageNet 2012 dataset with all the labeled training samples as the base categories, and we further select 5,000 categories from the ImageNet 2011 as novel categories. Concretely, 2,500 categories are used for validation and the rest 2,500 categories for final evaluation. Each category has  50 test samples.

For $k$-shot learning, only $k$ labeled images of the novel categories are used. We follow previous works \cite{hariharan2017low,wang2018low} to repeat the process by $5$ times and report the average accuracy. Here, we set $k$ as 1, 2, 5, 10 on ImageNet-FS and ImageNet-6K.

\noindent\textbf{Evaluation Metrics. }
We follow previous works \cite{wang2018low} to evaluate our proposed model on the top-5 accuracy of the novel categories (\textbf{Acc of novel}) and all (base + novel) categories (\textbf{Acc of all}).

\subsection{Comparison with State-of-the-Art}

\noindent\textbf{Performance on ImageNet-FS dataset. }
We present the results of the above metrics on 1, 2, 5, 10 shot learning on the ImageNet-FS in Table \ref{table:imagenet-sota}. As shown, our proposed model outperforms all existing methods by a sizable margin. Take the ``novel'' metric as example, our model achieves the accuracies of 62.1\%, 70.9\%, 78.4\%, and 82.3\%, outperforming current best results by 2.0\%, 1.3\%, 1.0\%, and 0.3\%, on 1, 2, 5, 10 shot learning, respectively.

Notably, compared with KTCH which also introduces word embedding as the external knowledge, we construct semantic correlation based on word embedding to transferring information between classifier weight and obtain an improvement of 4.0\%, 3.6\% in 1-shot and 2-shot setting, which shows our superiority of utilizing external knowledge. Besides, it is worth noting that the improvement is more notable if the samples of novel categories are fewer, e.g., 7.4\% on 1-shot learning v.s. 0.3\% on 10-shot learning. One possible reason for this phenomenon is that learning from fewer samples is more challenging and thus depends more on prior knowledge. Similar improvements are observed on the ``all'' metrics.

\noindent\textbf{Performance on ImageNet-6K dataset. }
All the foregoing comparisons focus on recognizing about 500 categories. In fact, there are much more categories in a real-world setting. To evaluate the proposed model on larger scale categories, we further conduct experiments on the ImageNet-6K dataset. As SGM \cite{hariharan2017low} and AWG \cite{gidaris2018dynamic} are current leading methods, we use the released codes to implement these methods on this dataset for comparison. Following \cite{hariharan2017low}, we train models with and without generating samples for the novel categories. The comparison results are presented in Table \ref{table:imagenet-sota}. Even though ImageNet-6K is more challenging and much larger in category size, our proposed model still outperforms existing methods. Specifically, it obtains the ``novel'' accuracy of 30.5\%, 37.5\%, an improvement of 2.9\%, 1.6\% on the 1-shot and 2-shot learning compared with existing methods. This comparison suggests the proposed method is scalable in category size.

\subsection{Ablative study}
\noindent\textbf{Analysis of Knowledge Embedding. }
The core of our proposed model is the prior knowledge graph that correlates base and novel categories to help regularize parameter propagation. As discussed above, our model follows SGM \cite{hariharan2017low} to use ResNet-50 as feature extractor and also use identical loss function for optimization, thus SGM can be regarded as the baseline of our model. In this part, we emphasize the comparison with SGM to evaluate the effectiveness of knowledge embedding model. As shown in Table \ref{table:imagenet-sota}, our framework significantly outperforms SGM, with accuracy improvement of 7.8\%, 7.6\% on two metrics in 1-shot setting.

To further analyze the effect of knowledge embedding, we further replace the category correlations with other non-informative form to demonstrate its benefit. Specifically, we consider the following two variants: 1) Uniform graph in which all the correlation values are uniformly set as $\frac{1}{K}$ and 2) Random graph in which all the correlation values are randomly selected from a uniform distribution. The comparison results are presented in Table \ref{table:knowledge}. We find that these two variants perform comparably with each other as both incur no additional information. Note that they achieve slightly better results than the baseline SGM. One possible reason is that knowledge propagation can help to better learn the classifier weights of novel categories. Still, our model with prior knowledge embedding significantly outperforms both two variants on all metrics for 1, 2, 5, 10 shot learning. These comparisons clearly demonstrate the benefit of knowledge embedding.

\renewcommand{\arraystretch}{1.2}
\begin{table}[t]

  \centering
  \fontsize{7}{7}\selectfont
    \begin{tabular}{cccccccccc}
    \toprule
    \multirow{2}{*}{ Graph}&
    \multicolumn{4}{c}{ Novel}&\multicolumn{1}{c}{ }&\multicolumn{4}{c}{ All}\cr
    \cmidrule(lr){2-5} \cmidrule(lr){7-10}
    &1&2&5&10& &1&2&5&10\cr
    \midrule
   u-graph     & 53.4      &67.4       &77.8       &81.5       & &63.8       &73.3       &80.3       &82.9\cr
   r-graph  & 54.4      &67.4       &77.8       &81.9       & &64.5       &73.3       &80.5       &83.2\cr
   c-graph   & 60.1      &69.4       &78.1       &82.1       & &67.0       &74.4       &80.7       &83.3\cr
   s-graph     &{\bf{62.1}}&{\bf{70.9}}&{\bf{78.4}}&{\bf{82.3}}& &{\bf{68.3}}&{\bf{75.2}}&{\bf{80.8}}&\bf{83.5}\cr
    \bottomrule
    \end{tabular}
    \caption{Top5-accuracy in ``novel'' and ``all'' metrics of our proposed model with semantic correlation knowledge (s-graph), with category hierarchy knowledge (c-graph), uniform correlation value (u-graph), and random correlation value (r-graph).}
  \label{table:knowledge}
\end{table}

\noindent\textbf{Analysis on Different Kinds of Knowledge. }
The correlations between categories can be constructed based on different kinds of knowledge, e.g., semantic similarity and category hierarchy in the paper. Here, we further conduct experiments with these two kinds of knowledge and present their results in Table \ref{table:knowledge}. We find that introducing both kinds of knowledge leads to obvious improvement than the baseline SGM and those with non-informative correlations, which suggests that our model can adapt to different kinds of knowledge. On the other hand, we find introducing semantic similarity knowledge achieves slightly better performance than introducing category hierarchy. One possible reason is that semantic similarity provides stronger and more direct correlations among different categories.

\noindent\textbf{Analysis on Different Similarity Metrics. }
In this work, the classifier is viewed as similarity measure between the input feature and the classifier weight, and three different similarity metrics, including inner product, cosine similarity and Pearson correlation coefficient, are evaluated.  As shown in Table \ref{table:imagenet-sota}, with different metrics, we still achieve better performance than the previous best result. Besides, we notice that inner product leads to slightly better accuracy than other metrics on ImageNet-FS, while cosine similarity and Pearson correlation coefficient perform better in much larger scale scenarios, for example, cosine similarity achieves higher accuracy than the inner product, with an increase of 1.0\% and 0.7\% for 1-shot and 2-shot setting in ImageNet-6K.

\begin{figure}[!t]
   \centering
   \includegraphics[width=1.0\linewidth]{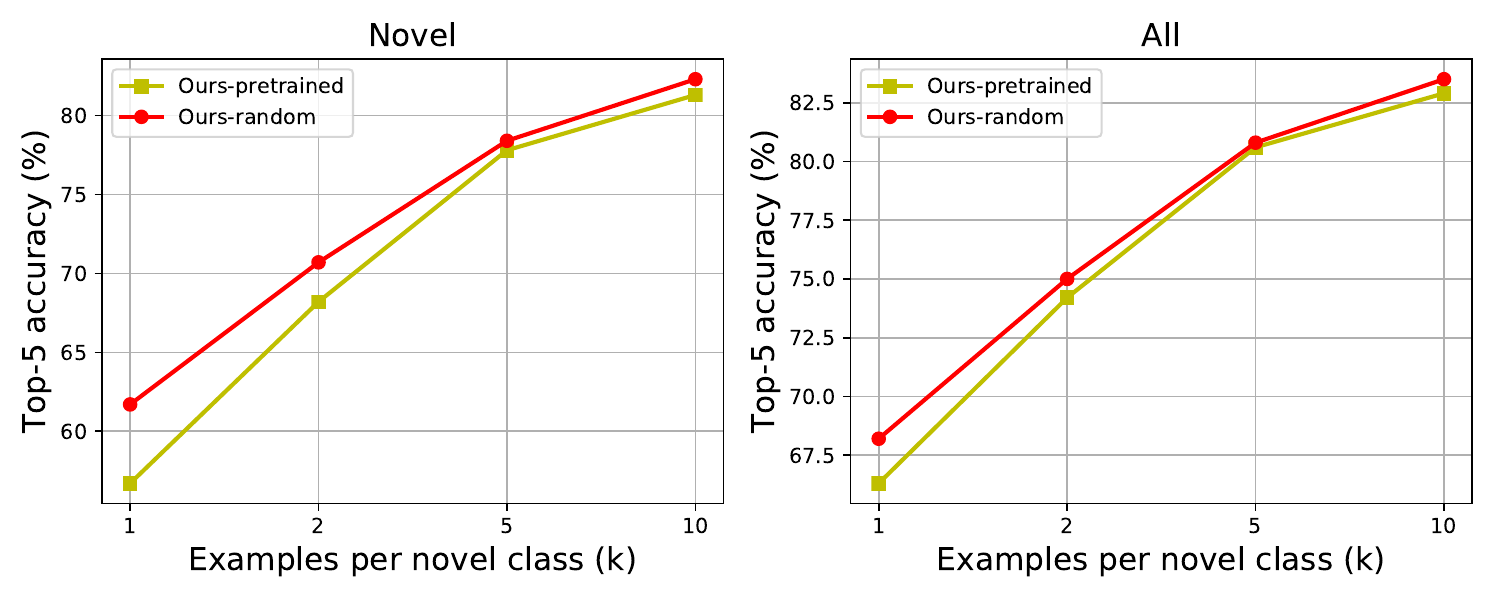} 
   \caption{Top5-accuracy in ``novel'' and ``all'' metrics of our model that initializes $\mathbf{W}_{base}$ with random values (Ours-random) and with the parameters pre-trained at the first training stage (Ours-pretrained).}
  \label{fig:initialization}
\end{figure}

\noindent\textbf{Analysis on Classifier Weight Initialization. }
At the second stage of the training process, we randomly initialize both $\mathbf{W}_{base}=\{\mathbf{w}_k\}_{k=1}^{K_{base}}$ and $\mathbf{W}_{novel}=\{\mathbf{w}_k\}_{k=K_{base}+1}^{K}$ to initialize the graph nodes. Actually, it is also intuitive to initialize $\mathbf{W}_{base}$ with the parameters pre-trained at the first stage and initialize $\mathbf{W}_{novel}$ randomly. Here, we further compare these two initialization settings and present the results in Figure \ref{fig:initialization}. We find that random initialization leads to much better results, up to 5.0\% accuracy improvement on 1-shot learning for the novel categories. The reason for this phenomenon is updating $\mathbf{W}_{base}$ from scratch enables training the graph propagation mechanism using sufficient samples of the base categories and can be easily generalized to update the weights of the novel categories.

\renewcommand{\arraystretch}{1.2}
\begin{table}[t]
  \centering
  \fontsize{8}{8}\selectfont
  \setlength{\tabcolsep}{1.5mm}{

    \begin{tabular}{lccccc}
    \hline
    \multirow{2}{*}{Method}&
    \multicolumn{5}{c}{Novel}\cr
    \cmidrule{2-6}
    &1&2&3&4&5\cr
    \midrule
    \hline
    \emph{\textbf{ImageNet2012/2010}}\cr
    NN           &34.2      &43.6       &48.7       &52.3       &54.0 \cr
    SGM\cite{hariharan2017low}          &31.6      &42.5       &49.0       &53.5       &56.8 \cr
    PPA\cite{qiao2018few}         &33.0      &43.1       &48.5       &52.5       &55.4 \cr
    LSD\cite{douze2018low}          &33.2      &44.7       &50.2       &53.4       &57.6 \cr
    KTCH\cite{li2019large}         &\textit{39.0}      &\textit{48.9}       &\textit{54.9}       &\bf58.7       &\textit{60.5} \cr
    \hline
    Ours    &\bf42.5&{\bf{50.3}}&{\bf{55.4}}&\textit{58.4}&\bf60.7\cr
    \hline

    \end{tabular}}
    \vspace{2pt}
    \caption{Top5-accuracy on ImageNet2012/2010 dataset.}
  \label{table:imagenet2012/2010-sota}
\end{table}

\subsection{Further evaluation}
\noindent\textbf{Comparison on the large scale few-shot benchmark ImageNet2012/2010. }
Here we also evaluate our method on another large scale few-shot benchmark \textbf{ImageNet2012/2010}, proposed by \cite{li2019large}. Briefly, in ImageNet2012/2010, all 1,000 categories from ILSVRC2012 are considered as the base categories, and 360 novel categories from ILSVRC2010, which are not overlapped with the base categories, are used as the novel categories. The base categories cover 200,000 labeled samples and the novel categories have 150 test samples. Same with \cite{li2019large}, We set $k$ as 1, 2, 3, 4, 5. Since we are not able to use the same training samples as \cite{li2019large} did, we randomly select training samples for $5$ times as we did on ImageNet-FS dataset and report the mean accuracy as the final result.

We compare with several methods on this benchmark including NN (nearest neighbor) \cite{li2019large}, SGM \cite{hariharan2017low}, PPA (parameter prediction from activations) \cite{qiao2017few}, LSD (large-scale diffusion) \cite{douze2018low}, and KTCH \cite{li2019large}. The comparative results are shown in Table \ref{table:imagenet2012/2010-sota}. It can be seen that our method achieves superior performance in most settings and exceeds over the best reported results by 3.5\%, 1.4\%, 0.5\% in 1, 2, 3 shot. Notably, KTCH also incorporates semantic word embedding as prior knowledge. Both KTCH and our method achieve significant improvements over other competing methods. Moreover, our method obtains superior results than KTCH, demonstrating its effectiveness.

\section{Conclusion}
In this work, we explore incorporating prior knowledge of category correlations to guide exploiting knowledge from base categories to help learn the classifier weights of novel categories. To this end, we formulate a novel Knowledge Graph Transfer Network model, which correlates classifier weights with a graph constructed based on their correlations and introduce a graph neural network to iteratively propagate and update classifier weights. Extensive experiments on the widely used ImageNet-FS and newly constructed ImageNet-6K dataset demonstrate the effectiveness of our proposed model over state-of-the-art methods.

\small
\bibliographystyle{aaai}
\bibliography{reference}

\end{document}